\documentclass[10pt,letterpaper]{article}

\usepackage{cogsci}

\cogscifinalcopy 

\usepackage{pslatex}
\usepackage{apacite}
\usepackage{float} 

\usepackage{natbib}
\usepackage{amsfonts}
\usepackage{amsmath}
\usepackage{graphicx}
\usepackage{color}
\usepackage{caption}
\usepackage{subcaption}
\usepackage{multicol}

\DeclareMathOperator*{\argmax}{argmax}

\title{Resource-rational Task Decomposition  \\to Minimize Planning Costs}

\author{
    {\large \bf 
        Carlos G. Correa$^*$
        (cgcorrea@princeton.edu) 
    } \\
    Princeton Neuroscience Institute, 
    Princeton University, 
    Princeton, NJ 08542 USA \\
\AND 
  {\large \bf
        Mark K. Ho$^*$
        (mho@princeton.edu)
    } \\
  {\large \bf
       Fred Callaway 
        (fredcallaway@princeton.edu)
    } \\
  {\large \bf
        Thomas L. Griffiths
        (tomg@princeton.edu)
    } \\
    Department of Psychology, 
    Princeton University, 
    Princeton, NJ 08542 USA \\
    $^*$Contributed equally
}

\begin{document}

\maketitle

\begin{abstract}

People often plan hierarchically. That is, rather than planning over a monolithic representation of a task, they decompose the task into simpler subtasks and then plan to accomplish those. Although much work explores \emph{how} people decompose tasks, there is less analysis of \emph{why} people decompose tasks in the way they do. Here, we address this question by formalizing task decomposition as a \textit{resource-rational representation problem}. Specifically, we propose that people decompose tasks in a manner that facilitates efficient use of limited cognitive resources given the structure of the environment and their own planning algorithms. Using this model, we replicate several existing findings. Our account provides a normative explanation for how people identify subtasks as well as a framework for studying how people reason, plan, and act using resource-rational representations.

\end{abstract}

\section{Introduction}
Your brother's birthday is coming up, so you decide to leave work early to drop his gift off at the post office. Although you know your way around town, you haven't been to the post office in several months, so you need to think. In particular, you start to \emph{plan}: ``How do I get to the post office from here?'' you ask yourself. ``Well, there's that caf\'{e} where I sometimes get my morning coffee. If I can first get there, then I should be able to get to the post office easily. Now, how should I get to the caf\'{e}? I know its east of where I am, so I can walk that way until I hit the main road...'' Continuing this line of thought for a few moments, you come up with a plan before setting off with determination.

The seemingly mundane choice to navigate to the caf\'e before navigating from the caf\'e to the post office is an example of \emph{task decomposition}. That is, rather than reason about a task in its totality (e.g., going from work to the post office), people decompose a task into manageable subtasks (e.g., going from work to the caf\'e; going from the caf\'e to the post office) and then reason in terms of those subtasks. Planning at multiple levels of abstraction has been extensively documented in psychology and neuroscience~\citep{botvinick2009} and plays an important role in developing systems that can solve complex, high-dimensional problems~\citep{sacerdoti1974planning}. In short, hierarchical planning and decision-making is a key element of intelligent behavior in both humans and machines.

Although much research has explored how people leverage hierarchical representations~\citep{ribasfernandes2011neural,Cushman13817,balaguer2016neural}, there has been less systematic investigation into the principles that determine task decompositions in the first place. There are a few notable exceptions, including accounts that emphasize the value of inferring the hidden structure to guide behavior across tasks~\citep{Collins2013,Tomov2020} as well as accounts based on compressing a representation of optimal behavior~\citep{solway2014,Maisto2015}. But, whereas these existing models emphasize decomposition in relation to statistical inference about the environment or behavior, our account focuses on a separate role that task decomposition plays: It makes \emph{reasoning} easier.

Here, we approach task decomposition as a \emph{resource-rational representation problem}. That is, we model people as solving the problem of how to break down a task in a manner that makes efficient use of planning resources. In the following sections, we provide background on related work before discussing the mathematical details of our normative account of task decomposition. We then report several simulations and show how our model can explain human data from four experiments reported by~\cite{solway2014}. Finally, we conclude by discussing future directions for resource-rational approaches to problem solving representations.

\section{Background}

Planning is hard because of the \emph{curse of dimensionality}~\citep{bellman1957dynamic}: As one attempts to plan into an increasingly distant future, over a larger state space, or under conditions of greater uncertainty, computation quickly becomes intractable. Nonetheless, humans have numerous strategies that allow us to plan in complex domains. Some of these strategies involve modifying the \emph{search process}. For example, during search, people have been shown to limit their depth of planning~\citep{MacGregor2001,keramati2016}, prune away unpromising paths~\citep{Huys2012}, and direct their search using model-free value estimates~\citep{Anderson1990,NewellSimon1972,vanopheusden2017}. Another strategy is to modify the \emph{problem representation} itself. Various forms of hierarchical planning~\citep{botvinick2012hierarchical} and task decomposition~\citep{solway2014,Huys2015} are characteristic of this approach. But while these two types of strategies are distinct, they are also clearly intertwined: How one represents a problem can make search anywhere from impossible to trivial~\citep{kaplan1990search}.  

The present work takes inspiration from the deep relationship between search---a type of computation---and task decomposition---a type of representation---in the cognitively demanding setting of planning. Task representations can play a key role in making problem-solving computations more efficient~\citep{ho2019}, and identifying general principles for automatically learning such representations is an active area of research in artificial intelligence. For instance, \cite{jinnai2018} examine how penalizing dynamic programming iterations can guide decomposition, while \cite{harb2018} introduce a deliberation cost for switching subtasks to help shape a decomposition. Here, we extend these ideas by analyzing human task decomposition in terms of \emph{search costs}. Broadly, our approach is in the spirit of \emph{resource-rational analysis}~\citep{Griffiths2015,Lieder2020}, a formal framework for deriving cognitive models under the assumption that people make rational use of their limited cognitive resources. Previous resource-rational analyses of planning have focused primarily on the search process itself~\citep[e.g.,][]{cogsci18-Callaway}. However, the interdependence of computations and the representations over which they operate means that this general framework can be readily applied to the latter.

\begin{figure*}[ht]
\centering

\begin{multicols}{4}
\begin{subfigure}[b]{.2\textwidth}
    \hspace*{1.4em}
    \includegraphics[width=.9\textwidth]{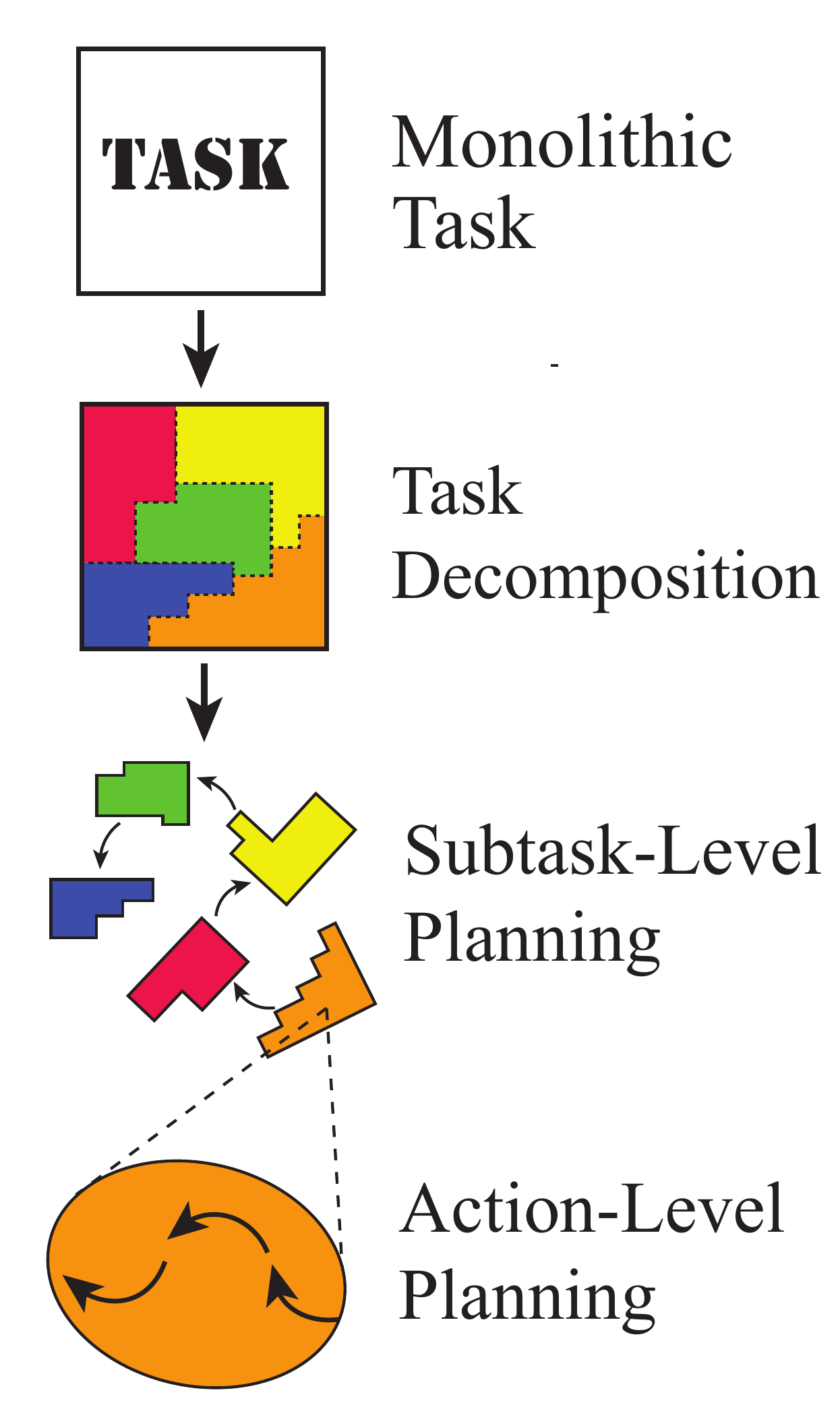}
    \caption{}
\label{fig:model_levels}
\end{subfigure}
\begin{subfigure}[b]{.25\textwidth}
  \includegraphics[width=\textwidth]{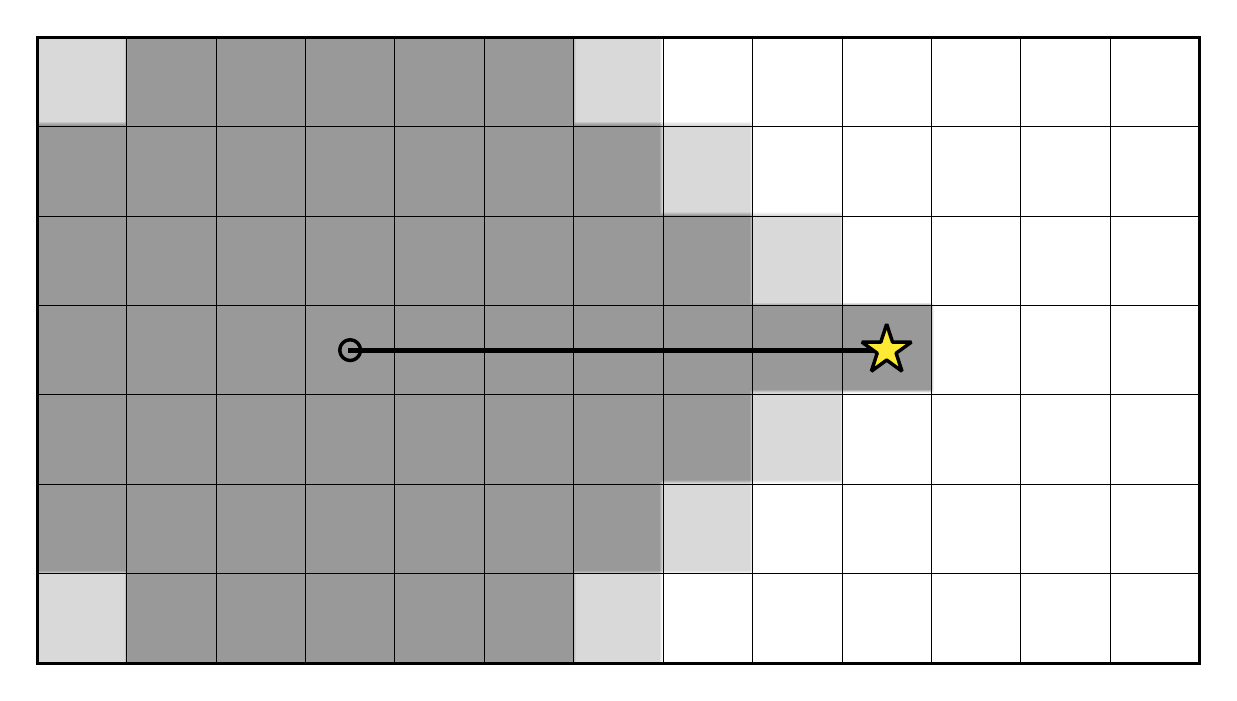}
  \label{fig:field_bfs_no_option}
\end{subfigure}
\begin{subfigure}[b]{.25\textwidth}
  \centering
  \includegraphics[width=\textwidth]{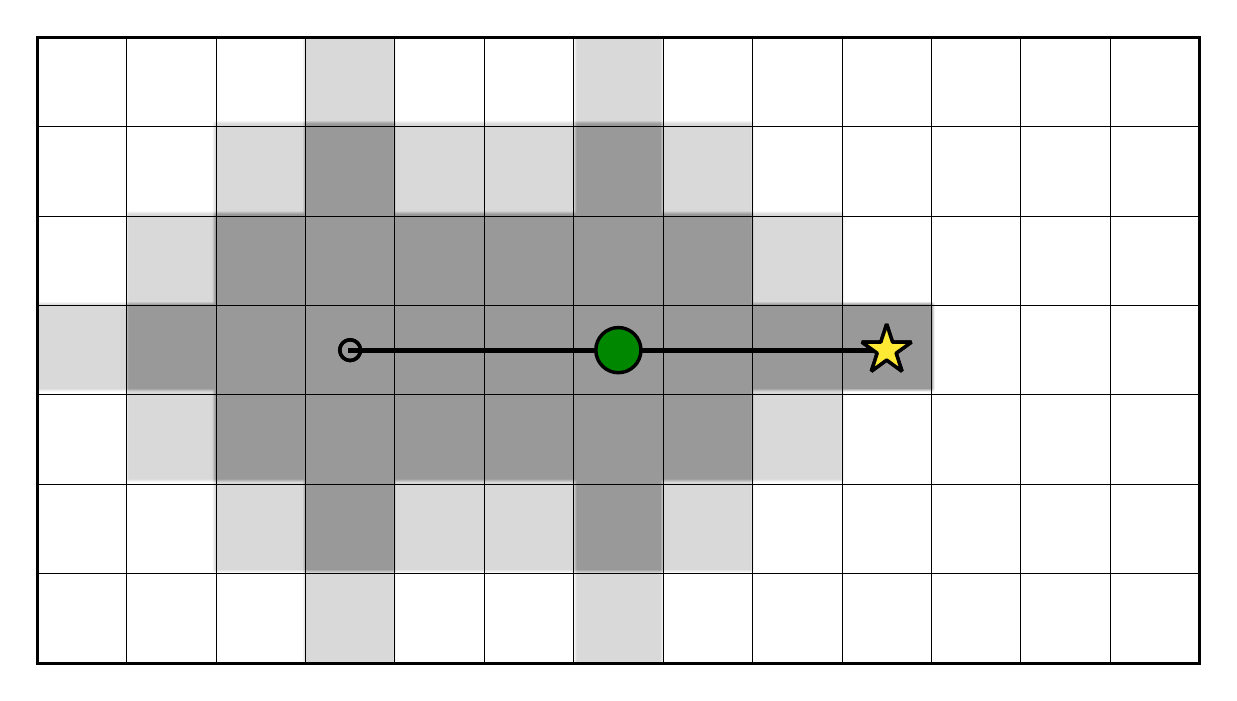}
    \caption{Breadth-First Search}
  \label{fig:field_bfs_option}
\end{subfigure}
\begin{subfigure}[b]{.25\textwidth}
  \includegraphics[width=\textwidth]{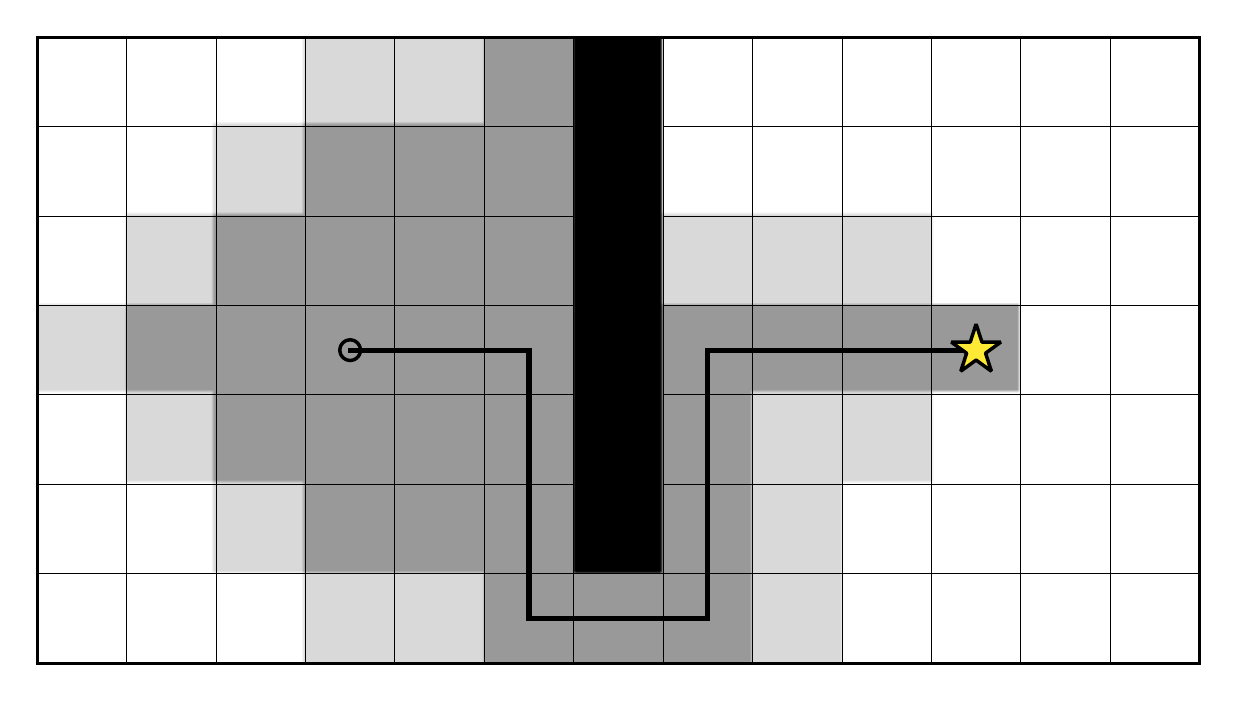}
  \label{fig:tworoom_astar_nooptions}
\end{subfigure}
\begin{subfigure}[b]{.25\textwidth}
  \centering
  \includegraphics[width=\textwidth]{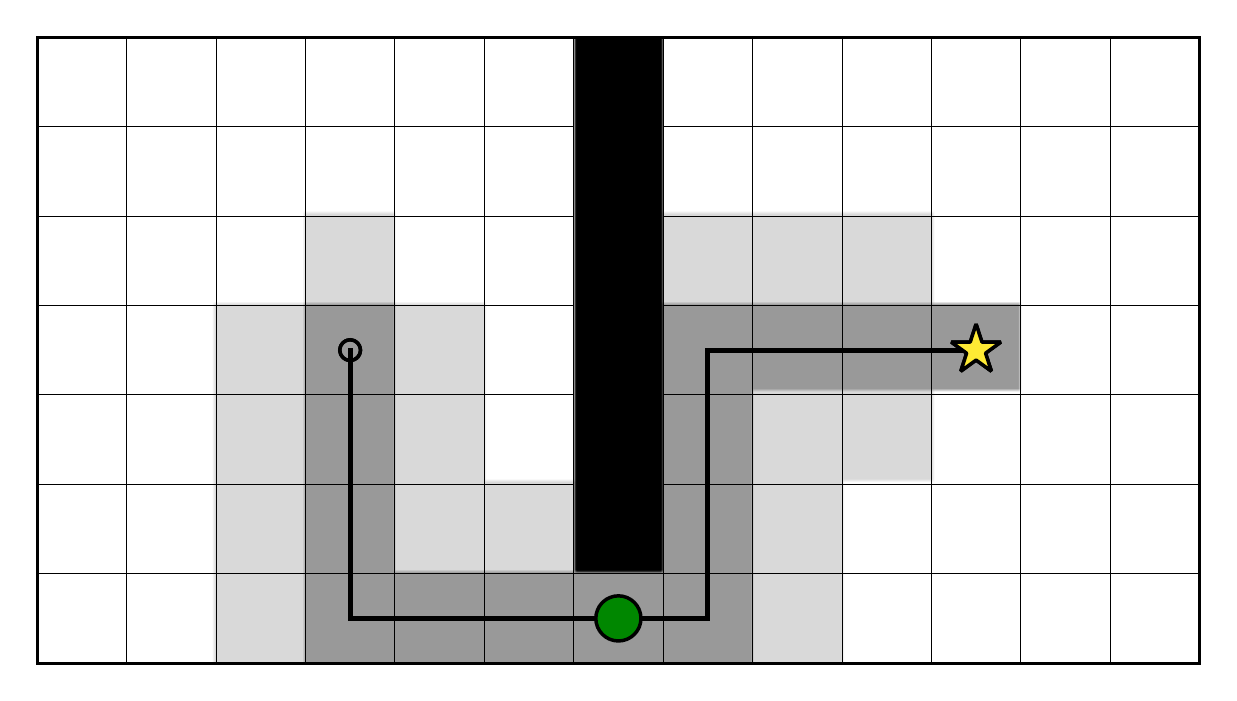}
    \caption{A$^*$ Search}
  \label{fig:tworoom_astar_options}
\end{subfigure}
\begin{subfigure}[b]{.14\textwidth}
  \includegraphics[width=\textwidth]{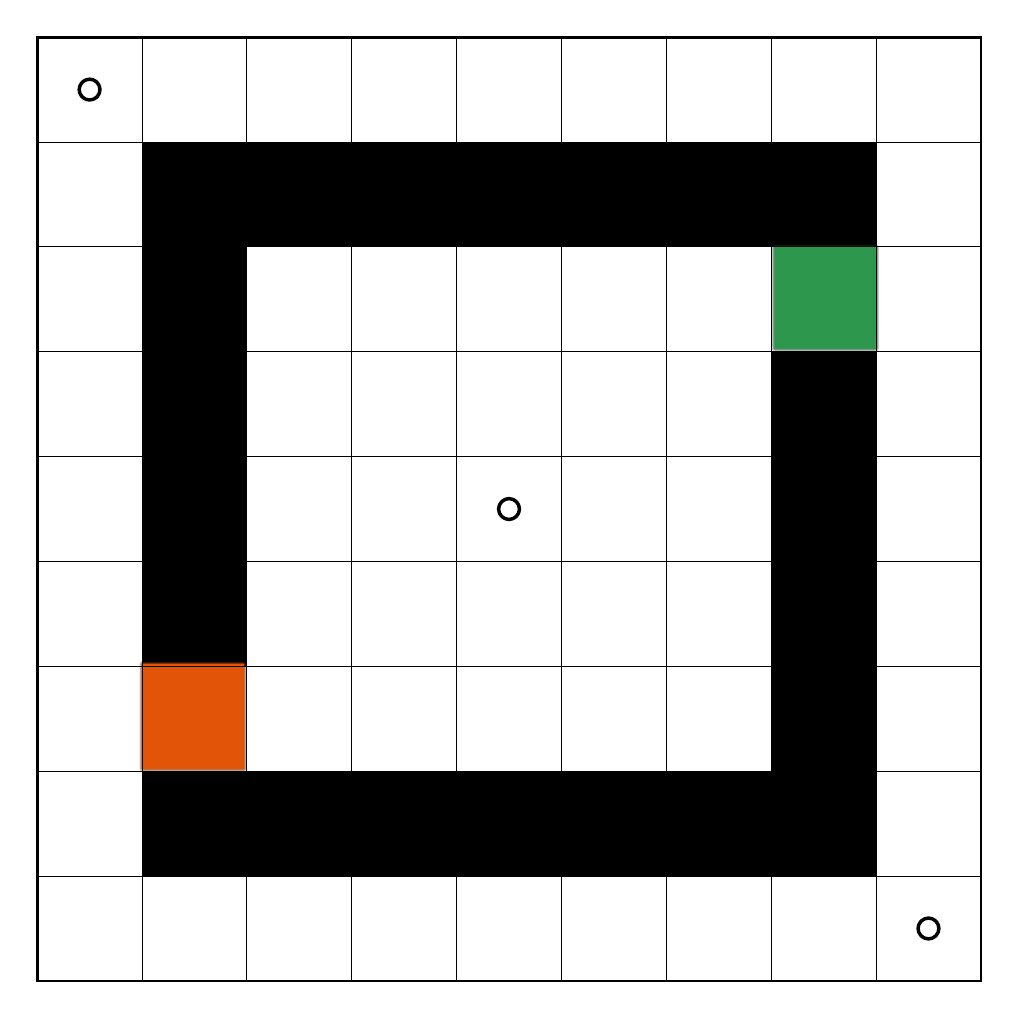}
  \label{fig:indooroutdoor_options}
\end{subfigure}
\begin{subfigure}[b]{.14\textwidth}
  \centering
  \includegraphics[width=\textwidth]{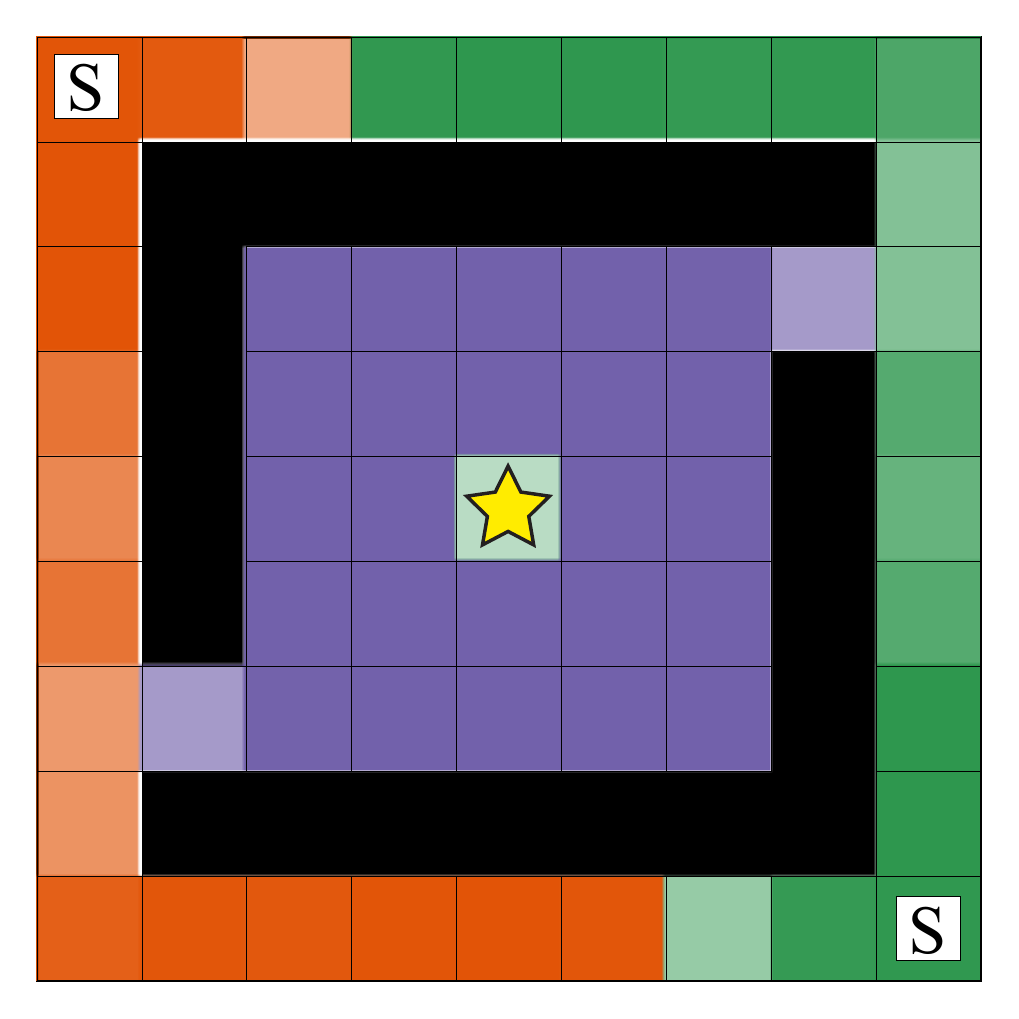}
    \caption{}
  \label{fig:indooroutdoor_policy}
\end{subfigure}
\end{multicols}
\vspace{-6mm}
\caption{(A) Our account relates three levels of optimization during problem solving: task decomposition, subtask-level planning, and action-level planning. (B) \emph{Top}: Planning with breadth-first search (BFS) from a start state (small circle) to the goal state (star) with no subgoals. Grey squares indicate nodes visited during search process (49 nodes visited). \emph{Bottom}: BFS with the optimal size $1$ task decomposition, where the green circle represents a subgoal (26 nodes visited). (C) \emph{Top}: Planning with A$^*$ search with a Manhattan distance heuristic and no subgoals (28 nodes visited). \emph{Bottom}: A$^*$ search with the optimal size $1$ task decomposition (13 nodes visited). (D) Indoor/Outdoor domain with optimal size $2$ task decomposition. \emph{Top}: The three dots represent possible start/goal states. The green and red tiles are the non-trivial subgoals in the optimal task decomposition. \emph{Bottom}: Subtask-level policy when the center tile is the goal (star) and the diagonal corners are possible start states (``S''). Colors correspond to the most likely subgoal chosen at each state. Purple is the trivial ``go to goal'' subtask.}
\label{fig:gridworlds}
\end{figure*}

\section{Resource-Rational Task Decomposition}
By treating task decomposition as a resource-rational problem, we assume people acquire task representations that enable them to plan efficiently and perform tasks successfully. Our account distinguishes between three nested levels of optimization (Figure~\ref{fig:model_levels}). The lowest level is \emph{action-level planning}, where concrete actions are chosen that solve a subtask (e.g., which direction should I walk to get to the caf\'e). The next level is \emph{subtask-level planning}, where a sequence of subtasks is chosen (e.g., navigating to the caf\'e and then to the post office). Finally, the highest level is \emph{task decomposition}, where a set of subtasks that constitute the decomposed task is chosen (e.g., setting the caf\'e as a possible subgoal across multiple tasks). 

Importantly, solutions to the higher levels of optimization depend on what happens at lower levels: A good task decomposition depends on how the subtask-level planner will compose the subtasks, and the selection of subtasks depends on how the action-level planner will accomplish each one. Furthermore, a resource-rational task decomposition is sensitive not only to how well the planners solve their subproblems (e.g., does action-level planning identify a good route to the caf\'e?), but also on the computational cost of identifying those solutions (e.g. how much thought did it take to find that route?). Next, we discuss each of the three levels of our model.

\subsection{Action-level Planning}
Action-level planning computes the optimal actions that one should take to reach a subgoal. Here, we focus on deterministic, shortest path problems (e.g., finding a route to the caf\'e). Formally, action-level planning occurs over a task defined by a set of states, $\mathcal{S}$; a transition graph, $T \subseteq \mathcal{S} \times \mathcal{S}$; and a subgoal state, $z \in \mathcal{S}$. In our running example, states are possible locations (e.g., at the office, at work, at the caf\'e, at the post office); the transition graph represents which locations in town are accessible to one another; and a subgoal could be the caf\'e. 

Given an initial state, $s$, and a subgoal, $z$, action-level planning seeks to find a minimum-length sequence of states that begins at $s$ and ends at $z$. We denote the length of this minimum-length sequence to be $D(s, z)$. For computing the optimal sequence of actions, we consider two broad classes of search algorithms: uninformed search and heuristic search~\citep{NewellSimon1972,russell2009artificial}.

\subsubsection{Uninformed Search} When faced with a domain that lacks features to guide exploration, the best that a planning agent can do is blindly but systematically explore their model of the problem starting from an initial state. This strategy describes a broad class of search algorithms known as \emph{uninformed search}. For example, \emph{breadth-first search} (BFS) explores states in the order of their distance from the starting state. As a result, for an initial node $s$ and subgoal node $z$, BFS will explore all states that are less than the minimum path length $D(s, z)$, as shown in Figure~\ref{fig:field_bfs_option}. The cost of BFS, $C_{\texttt{BFS}}(s, z)$, is proportional to the number of these states.

\subsubsection{Heuristic Search} Unlike uninformed search, \emph{heuristic search} leverages domain knowledge in the form of a \emph{heuristic function} that can provide an optimistic estimate of the distance to a goal. For instance, when navigating to the caf\'e, you might know that it is North-East of work, leading you to consider walking North or East before South or West. The canonical heuristic search algorithm is A$^*$~\citep{hart1968formal}, which considers states in the order of an optimistic estimate of the total cost of a solution passing through that state (Figure~\ref{fig:tworoom_astar_options}). This estimate is the cost of reaching that state plus a lower bound on the cost from that state to the goal, which is given by the heuristic function. For example, when navigating to the caf\'e, one might use Euclidean distance as a heuristic, which is optimistic because it assumes you can walk directly towards your destination (e.g., no obstacles will be in the way). By prioritizing states that are more promising (as estimated by the heuristic), A$^*$ can search far fewer states than BFS, resulting in a lower search cost, $C_{\texttt{A}^*}(s, z)$.

\subsection{Subtask-Level Planning}
A number of formalisms have been used to model hierarchical decision-making~\citep{sutton1999,dietterich2000,parr1998}. Here, we assume a simple model of hierarchical planning that involves only a single level above action-level planning, which we call \emph{subtask-level planning}. Formally, subtask-level planning occurs over a set of subgoals, $\mathcal{Z} \subset \mathcal{S}$.\footnotemark Given a set of subgoals, subtask-level planning consists of choosing the best sequence of subgoals that accomplish a larger goal. Each subgoal is then provided to the action-level planner in turn, and the resulting action-level plans are combined into a complete plan to reach the goal state. For example, when navigating to the post office, the subtask-level planner might decide to first go to the caf\'e and then go to the post office from there, and the action-level planner would figure out the precise sequence of steps to get from work to the caf\'e and from the caf\'e to the post office.

\footnotetext{For readers familiar with the \emph{options framework}~\citep{sutton1999}, we note that what we call a subgoal is equivalent to a simple option where the set of initial states is the full state space, $\mathcal{S}$, and the termination function is $\beta(s) = \mathbf{1}(s = z)$. This means that subtask-level planning is a semi-Markov decision process.}

The objective of the subtask-level planner is to identify the sequence of subgoals that brings the agent to the goal state while maximizing task rewards and minimizing computational costs. Here, we focus on tasks in which the task is simply to reach the goal state in as few steps as possible. Additionally, note that we only consider action-level planners that return the optimal shortest path. Thus, formally, the task reward associated with choosing a subgoal $z$ from state $s$ is the negative distance: $R(s, z) = -D(s, z)$. We can then compactly express the optimization problem faced by the subtask-level planner as a Bellman equation \citep{bellman1957dynamic}. Given a task goal $g$, a set of subgoals $\mathcal{Z}$, and an algorithm with a search cost function $C_{\texttt{Alg}}$, the optimal subtask-level planning utility from any state $s \in \mathcal{S}$ is:
\begin{equation}
    V_{\mathcal{Z}}^g(s) = 
        \max_{z \in \mathcal{Z}} \left\{
            R(s, z) - C_{\texttt{Alg}}(s, z) + 
            V^g_{\mathcal{Z}}(z) 
            \right \}.
    \label{eq:subtask}
\end{equation}
The fixed point of Equation~\ref{eq:subtask} can be used to identify the optimal subtask-level policy~\citep{puterman1994markov}. Additionally, we assume that the ultimate goal, $g$, is always included in $\mathcal{Z}$ to ensure that it is possible for the subtask-level planner to solve the task. Finally, although we do not explore this possibility here, note that this formulation allows us to easily express tradeoffs between task rewards, $R$, and algorithm-specific computation costs, $C_{\texttt{Alg}}$.

\subsection{Task Decomposition}

Having defined action-level planning and subtask-level planning over subgoals, we can now turn to our original motivating question: How should people decompose tasks? In this context, this reduces to the problem of selecting the best set of subgoals. Importantly, we assume that people rely on a common set of subgoals for all the different possible tasks that they might have to accomplish in a given environment. 
For example, \textit{be at the downtown train station} is a good subgoal because it is often along relatively-optimal paths, whereas \textit{be at a friend's place on the other side of town} is probably not a good subgoal because it is only relevant when visiting that friend.

We formalize subgoal selection as an optimization problem: Identify the set of subgoals, $\mathcal{Z}^*$, that maximize the value attained by the subtask-level planner on average. That is,
\begin{equation}
    \mathcal{Z}^* = \argmax_{\mathcal{Z}}\mathbb{E}_{s, g} [V^g_{\mathcal{Z}}(s)],
\label{eq:constructor}
\end{equation}
where the expectation is with respect to a \emph{task distribution}, $p(s, g)$, over starting states $s$ and goals $g$. Importantly, this objective takes into account both the expected task rewards and the costs of action-level planning mediated by subtask-level planning. 

\begin{figure*}[!ht]
\centering
\begin{multicols}{4}
\begin{subfigure}{0.2\textwidth}
  \centering
  \includegraphics[height=.12\textheight]{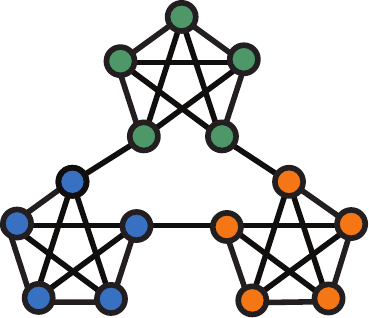}
\end{subfigure}
\begin{subfigure}{0.2\textwidth}
  \centering
  \includegraphics[height=.12\textheight]{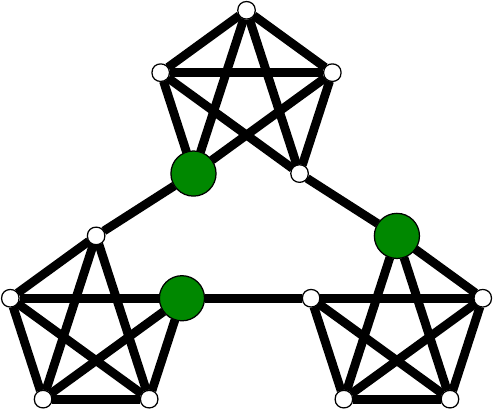}
  \caption{}
    \label{fig:shapiro}
\end{subfigure}

\begin{subfigure}{0.2\textwidth}
  \centering
  \includegraphics[height=.12\textheight]{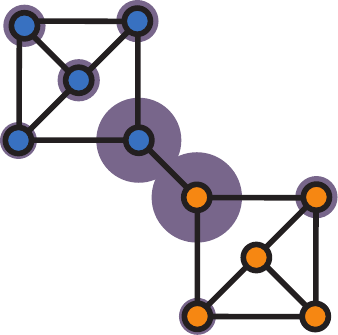}
\end{subfigure}
\begin{subfigure}{0.2\textwidth}
  \centering
  \includegraphics[height=.12\textheight]{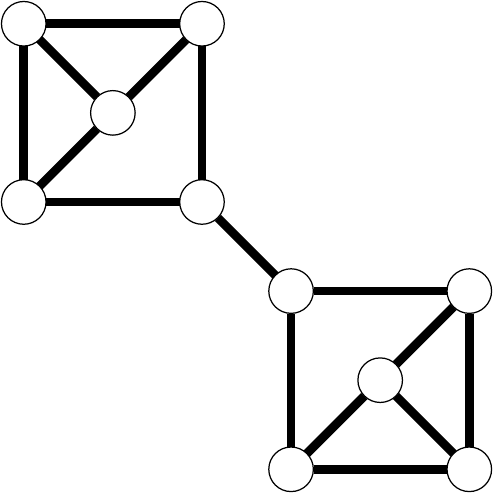}
    \caption{}
  \label{fig:solway_exp1}
\end{subfigure}

\begin{subfigure}{0.2\textwidth}
  \centering
  \includegraphics[height=.12\textheight]{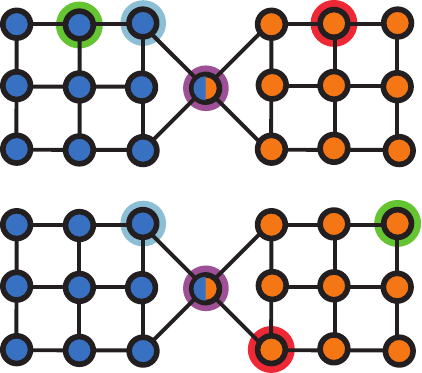}
\end{subfigure}
\begin{subfigure}{0.2\textwidth}
  \centering
  \vspace{2em}
  \includegraphics[width=\textwidth]{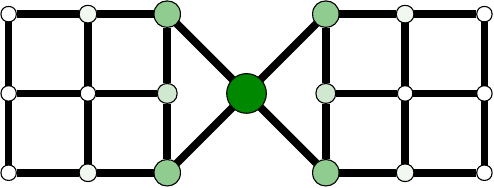}
  \vspace{1em}
\caption{}
  \label{fig:solway_exp2}
\end{subfigure}

\begin{subfigure}{0.2\textwidth}
  \centering
  \includegraphics[height=.12\textheight]{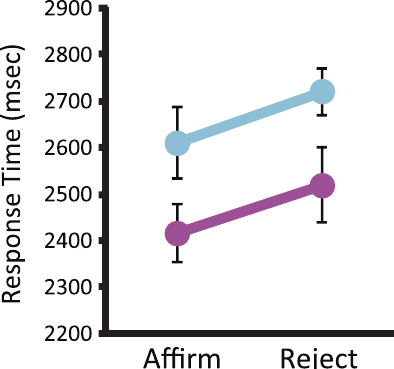}
\end{subfigure}
\begin{subfigure}{0.2\textwidth}
  \centering
  \includegraphics[height=.12\textheight]{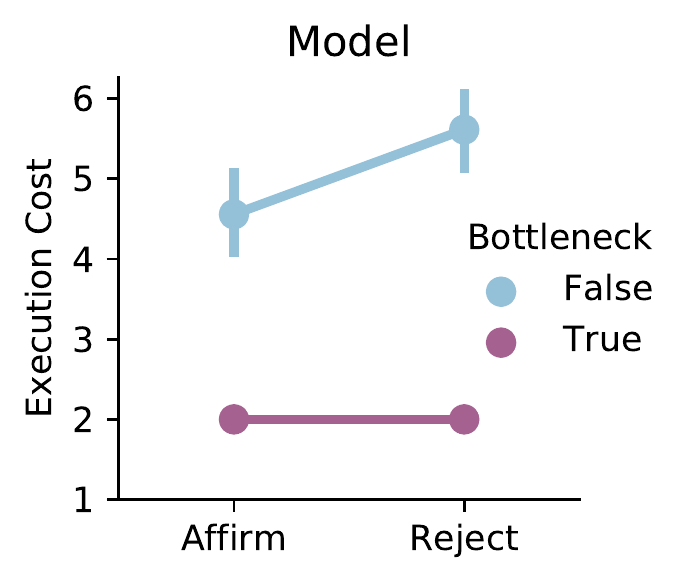}
  \caption{}
  \label{fig:solway_exp3}
\end{subfigure}

\end{multicols}

\caption{Analyzing \protect\cite{solway2014} Experiments 1-3 with uninformed search. (A) \emph{Top}: \protect\cite{solway2014} decomposition of graph from \textbf{\protect\cite{Schapiro2013}}. \emph{Bottom}: Resource-rational task decomposition. (B) \emph{Top}: Graph used in \textbf{Experiment 1} from \protect\citeauthor{solway2014} with proportion of ``bus stops'' placed at states. \emph{Bottom}: Subgoals do not facilitate more efficient breadth-first search (BFS) in this graph, so our model does not learn to use any states as subgoals. (C) \emph{Top}: Graph for \textbf{Experiment 2} with bottlenecks depicted. Same graph was used in Experiment 3. \emph{Bottom}: Distribution over subgoals $z$ from softmax of value of subgoal with inverse temperature of 100. Task decomposition facilitates more efficient BFS in this graph. In particular, the ``bottleneck state'' connecting the two regions is the optimal subgoal, followed by the adjacent states. (D) \emph{Top}: \textbf{Experiment 3} in \protect\citeauthor{solway2014} probed whether participants' plans included states in the graph. Responses were significantly faster for bottleneck (purple) compared to non-bottleneck states (blue). \emph{Bottom}: Simulations of probe response computations in experimental trials with the optimal task decomposition learned by our model. Our model takes fewer steps to respond when the probe is a bottleneck state, mirroring the experimental findings.
}
\label{fig:solway_comparison}
\end{figure*}

\subsection{Implementation}
The code for all the analyses we report is available at https://bit.ly/2T44Tun. Here, we briefly describe the implementation. For both BFS and A$^*$, we calculated action-level computational costs $C_{\texttt{Alg}}(s, z)$ and minimum path lengths $D(s, z)$ for every state $s \in \mathcal{S}$ and subgoal $z \in \mathcal{S}$. With these quantities, a set of subgoals $\mathcal{Z}$, and distribution over goals and starting states $p(s, g)$, we can define the optimal expected subtask-level planning value function, $\mathbb{E}_{s, g}\left[V^g_{\mathcal{Z}}(s)\right]$ (see Equation~\ref{eq:subtask}). We compute this function using value iteration with a threshold of $\varepsilon = 10^{-5}$~\citep{bellman1957dynamic}.

Finally, to solve for the optimal set of subgoals, $\mathcal{Z}^*$, we explored two methods. The first was an exact method---enumeration and evaluation of all subgoal sets. The second was a gradient-based method. This method used a differentiable version of value iteration at the subtask-planning stage~\citep{haarnoja2017,ho2020} and distributions over subgoals instead of discrete subgoals at the task decomposition level. 
While enumeration is intractable for large state spaces, we found that the methods produced similar results when both were feasible. Thus, we present results using the exact enumeration method when it was computationally tractable---for the small environments used in Solway et al. (2014)---and the gradient method otherwise.

\subsection{Gridworld Simulations}
To illustrate the properties of our model, we begin by analyzing optimal decompositions of simple gridworld tasks. The grids we tested include \emph{Open Field}, \emph{2-Room}, and \emph{Indoor/Outdoor}. We tested our model with both BFS and A$^*$. As shown in Figure~\ref{fig:gridworlds}, our model produces intuitive task decompositions as a function of a task and planning algorithm.

\section{Analysis of \protect\cite{solway2014} Experiments}

\begin{figure*}[ht]
\centering
\begin{subfigure}{0.2\textwidth}
  \centering
  \includegraphics[height=.12\textheight]{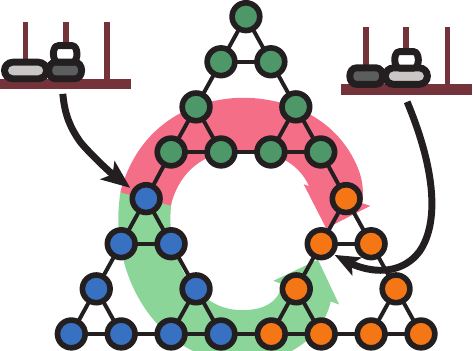}
  \caption{}
  \label{fig:solway_exp4}
\end{subfigure}
\begin{subfigure}{0.2\textwidth}
  \centering
  \includegraphics[height=.12\textheight]{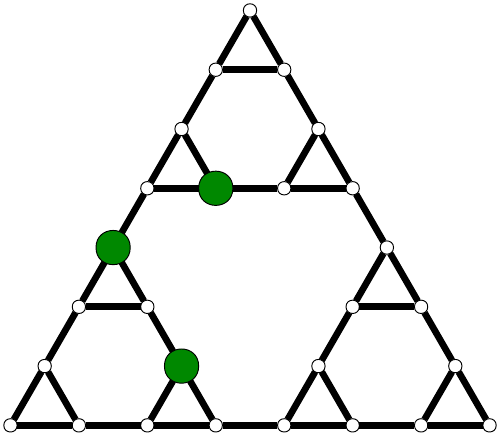}
  \caption{}
  \label{fig:td_hanoi_bfs}
\end{subfigure}
\hspace{-2.2em}
\begin{subfigure}{0.23\textwidth}
  \centering
  \vspace{-2.6em}
  \includegraphics[height=.195\textheight]{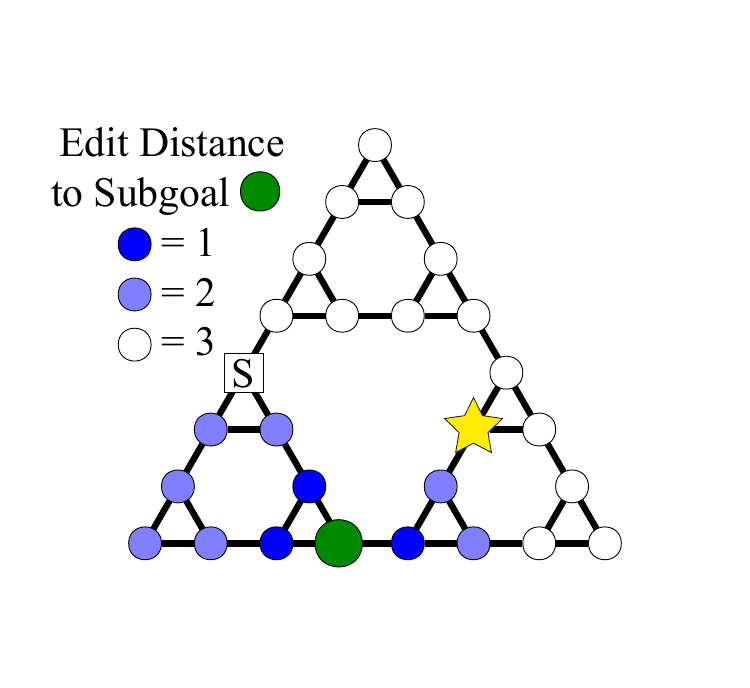}
  \vspace{-3.7em}
  \caption{}
  \label{fig:astar_heuristic}
\end{subfigure}
\begin{subfigure}{0.2\textwidth}
  \centering
  \includegraphics[height=.12\textheight]{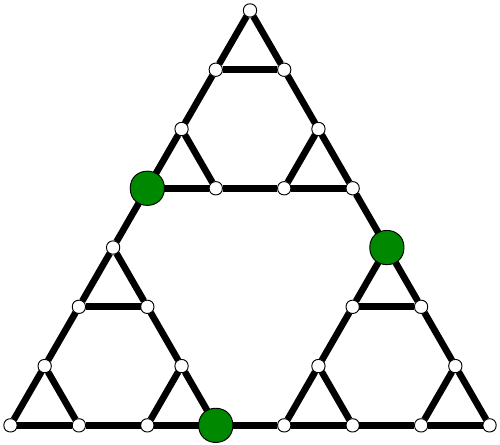}
  \caption{}
  \label{fig:td_hanoi_astar}
\end{subfigure}
\caption{Analyzing \protect\cite{solway2014} Experiment 4. (A) Hierarchically optimal (green) vs. non-hierarchically optimal (red) paths in Tower of Hanoi. (B) Optimal task decomposition with BFS has one bottleneck subgoal and two neighboring non-bottleneck states as subgoals. (C) Edit distance to a subgoal (green circle). Edit distance provides useful local information for heuristic search. (D) Optimal task decomposition with A* and edit distance heuristic has three equidistant bottleneck states as subgoals.}
\label{fig:hanoi}
\end{figure*}

\cite{solway2014} reported four studies that investigated how people decompose tasks and engage in hierarchical planning. Here, we ask if our resource-rational model can account for these findings. We first discuss Experiments 1-3 (Figure~\ref{fig:solway_comparison}), which relied on tasks in which people could not leverage prior knowledge and then turn to Experiment 4 (Figure~\ref{fig:hanoi}), which used the Tower of Hanoi~\citep{nilsson1971}, a task that allows for the use of prior knowledge.

\subsection{Experiments 1-3: Uninformed Search}

\subsubsection{Summary of Findings}
Experiments 1-3 reported by \cite{solway2014} experimentally tested the hierarchical structure used by participants when performing navigation tasks over abstract state spaces.
Figures~\ref{fig:solway_exp1} and \ref{fig:solway_exp2} show the connectivity structure of the domains people were given.

A key qualitative finding reported by \cite{solway2014} was that people's responses reflected a decomposition of the state space based on \textit{bottlenecks}, states or transitions that connect more densely-connected regions of a state space~\citep{csimcsek2009}. A preliminary modeling result they reported recovered a task decomposition along community boundaries in a graph (Figure~\ref{fig:shapiro}) studied in \cite{Schapiro2013}. Experiment 1 assessed this in the task with the transition structure in Figure~\ref{fig:solway_exp1} (top) by having participants choose a ``bus stop'' that would be most useful for making ``deliveries'' between locations represented by the icons. Participants overwhelmingly chose the bottleneck states, as displayed in the figure. Experiment 2 had participants actually navigate to and from random locations in the task with the structure in Figure~\ref{fig:solway_exp2}. However, on test trials, participants were asked to either identify locations along the path \emph{in any order} or identify a single location on the path. Participants tended to report bottleneck states first, suggesting they were thinking about these states first in their planning process. Finally, Experiment 3 used the same domain as Experiment 2, but participants were probed about whether a state was on the optimal path between two states. Participants answered faster for bottleneck states compared to non-bottleneck states, providing additional evidence that these states are the first to come to mind  (Figure~\ref{fig:solway_exp3}).

\subsubsection{Model Implementation and Results}
For the \citeauthor{Schapiro2013} graph (Figure~\ref{fig:shapiro}), we assumed a uniform distribution over all start and goal states, and set the number of subgoals to $|\mathcal{Z}| = 3$. The best subgoals separated the three communities at their boundaries, allowing the action-planner to first find the community containing the goal and then search for the goal within that community.

For the second graph (Figure~\ref{fig:solway_exp1}; \citeauthor{solway2014} Experiment 1), we again assumed a uniform distribution over all start and goal states, and set the number of subgoals to $|\mathcal{Z}| = 1$. Interestingly, no set of subgoals achieved greater value than planning without subgoals: $\mathcal{Z}^* = \emptyset$. Thus, the model did not reflect the empirical results. Although ``bus stop'' judgments recovered bottleneck states, they may reflect a process that is distinct from task decomposition for planning. Further experiments are needed to evaluate this.

We made the same assumptions for the third graph (Figure~\ref{fig:solway_exp2}; Experiments 2 \& 3), and found that the bottleneck state was always in the optimal decomposition. 
The next-best decompositions all included one of the four states connected to the bottleneck state,
indicating that navigating near the bottleneck state is a useful subgoal in this task.

To replicate the reaction time results reported in Experiment 3 (Figure~\ref{fig:solway_exp3}), we simulated hierarchical planning with the optimal decomposition using trials as described by~\citeauthor{solway2014} Specifically, the model first constructed a subtask-level plan. If the queried state was a subgoal state, the model replied ``Affirm'' as soon as the subtask-level planner encountered the state and ``Reject'' if the state was not in the completed subtask-level plan. If the queried state was not a subgoal state, the model proceeded to construct each action-level plan in turn. As soon as the queried state was encountered, the model replied ``Affirm''. If the final action-level plan to the goal was completed without encountering the state, the model replied ``Reject''. In either case, we used the total number of subgoals and states that were simulated before the response was produced as a proxy for reaction time.
These results are plotted for bottleneck vs. non-bottleneck probes and ``Affirm'' vs. ``Reject'' type probes in Figure~\ref{fig:solway_exp3}.

\subsection{Experiment 4: Tower of Hanoi and Heuristic Planning}

\subsubsection{Summary of Findings}
In a final experiment, \citeauthor{solway2014} tested participants solving the Tower of Hanoi~\citep{nilsson1971}. The experiment focused on ``problems of interest'', trials where two paths of the same length led to a goal but one crossed more bottleneck states. They found that participants preferred to take paths that crossed fewer bottleneck states. Assuming that people prefer hierarchically shorter paths (i.e., ones that use fewer subgoals), this has been taken to reflect a decomposition of the task based on bottleneck states.

\subsubsection{Model Implementation and Results}
The Tower of Hanoi is an important contrast to the tasks in the first three experiments because states have features that provide clues for search. For example, the \emph{edit distance} between two states provides an optimistic estimate of their minimum path length: it ignores that some transitions are forbidden and assumes you can rearrange blocks arbitrarily. Much like how spatial distance can guide planning in navigation tasks, heuristics like edit distance can guide problem solving in structured tasks.

To understand the relationships between heuristics, task decomposition, and \citeauthor{solway2014}'s results, we ran several versions of our model on the Tower of Hanoi. We set the number of subgoals to $|\mathcal{Z}| = 3$ and used BFS as the action-level planner. Notably, the optimal subgoals under this scheme were systematically ``skewed'', consisting of a bottleneck state and two nearby points (Figure~\ref{fig:td_hanoi_bfs}).
%
For our second simulation, we used the same procedure and parameters, but rather than using BFS (uninformed search), we used A$^*$ with an edit-distance heuristic for action-level planning (Figure~\ref{fig:astar_heuristic}). The top two decompositions both contained three bottleneck states in separate communities (Figure~\ref{fig:td_hanoi_astar}). Unlike BFS, A$^*$ can efficiently navigate between these points, allowing for a task decomposition that spans the full extent of the problem space.


\section{Discussion}
We have proposed a resource-rational account of task decomposition based on the idea that subgoals are decomposed to make planning easier.
Our model specifies three levels of nested optimization: \emph{Task decomposition} identifies a set of subgoals for a given domain, \emph{subtask-level planning} chooses sequences of subgoals to reach a goal, and \emph{action-level planning} chooses sequences of concrete actions to reach a subgoal.
Optimal task decomposition thus depends on both the structure of the environment and the computational resource usage specific to the planning algorithm.
We find that our model produces interpretable task decompositions in gridworld tasks and  decompositions consistent with three of the four findings reported by \cite{solway2014}.

The model presented here departs from and complements other normative proposals in the literature. Most existing approaches pose task decomposition as an \emph{inference problem}: People are modeled as inferring a generative model of the environment~\citep{Collins2013,Tomov2020} or as compressing optimal behavior~\citep{solway2014,Maisto2015}. In contrast, we pose task decomposition as a \emph{resource-rational representation problem}: People are modeled as having subgoals that reduce the computational overhead of action-level planning. This change in view has several consequences worth noting.

First, unlike inferential approaches that abstract away the underlying reasoning process, our framework requires specifying a planning algorithm. Different assumptions at this lowest level (e.g. using breadth first search or A*) can dramatically influence the task decomposition (e.g., Figure~\ref{fig:hanoi}). On the one hand, this makes model identification more challenging since the space of planning algorithms and parameterizations is vast. On the other hand, because our model is both algorithmic and normative, it can characterize the interplay of planning computations and representations in a well-posed manner.
Additionally, this approach allows us to analyze how behavioral suboptimality can arise from rational tradeoffs between task rewards and computation costs associated with particular search algorithms. Future empirical work on resource-rational planning representations will need to examine these questions in greater depth.

A second difference is that inferential models generally emphasize learning from task interactions as data, while we have deliberately set aside how resource-rational decompositions are learned. Specifically, our formulation assumes the existence of an optimization process that can select a decomposition, whether it be through direct experience with a task or other means. Although this temporarily defers important and interesting questions about online problem solving, characterizing any learning process requires first identifying what is being learned (i.e., what is being optimized). It is in this sense that the model presented here is a resource-rational analysis of task decomposition~\citep{Griffiths2015}.

More broadly, the work presented here is consistent with other recent efforts within cognitive science to understand how people engage in computationally efficient decision-making~\citep{Griffiths2015,lewis2014computational,Gershman2015computational,Lieder2020}. It is also complementary to recent work in artificial intelligence that explores the interaction between planning and task representations~\citep{jinnai2018,harb2018}. Our hope is that future work on human planning and problem solving will continue to investigate the relationships between computation, representation, and resource-rational decision-making.

\vspace{-1mm}
\section{Acknowledgements}
This work was supported by NSF grant \#1544924, AFOSR grant FA9550-18-1-0077 and grant 61454 from the John Templeton Foundation.

\vspace{-1mm}
\bibliographystyle{apacite}

\setlength{\bibleftmargin}{.125in}
\setlength{\bibindent}{-\bibleftmargin}

\bibliography{main}

\begin{thebibliography}{}

\bibitem [\protect \citeauthoryear {%
Anderson%
}{%
Anderson%
}{%
{\protect \APACyear {1990}}%
}]{%
Anderson1990}
\APACinsertmetastar {%
Anderson1990}%
\begin{APACrefauthors}%
Anderson, J\BPBI R.%
\end{APACrefauthors}%
\unskip\
\newblock
\APACrefYear{1990}.
\newblock
\APACrefbtitle {{The {A}daptive {C}haracter of {T}hought}} {{The {A}daptive
  {C}haracter of {T}hought}}.
\newblock
\APACaddressPublisher{Hillsdale, NJ}{Lawrence Erlbaum Associates, Inc.}
\PrintBackRefs{\CurrentBib}

\bibitem [\protect \citeauthoryear {%
Balaguer%
, Spiers%
, Hassabis%
\BCBL {}\ \BBA {} Summerfield%
}{%
Balaguer%
\ \protect \BOthers {.}}{%
{\protect \APACyear {2016}}%
}]{%
balaguer2016neural}
\APACinsertmetastar {%
balaguer2016neural}%
\begin{APACrefauthors}%
Balaguer, J.%
, Spiers, H.%
, Hassabis, D.%
\BCBL {}\ \BBA {} Summerfield, C.%
\end{APACrefauthors}%
\unskip\
\newblock
\APACrefYearMonthDay{2016}{}{}.
\newblock
{\BBOQ}\APACrefatitle {Neural Mechanisms of Hierarchical Planning in a Virtual
  Subway Network} {Neural mechanisms of hierarchical planning in a virtual
  subway network}.{\BBCQ}
\newblock
\APACjournalVolNumPages{Neuron}{90}{4}{893 - 903}.
\PrintBackRefs{\CurrentBib}

\bibitem [\protect \citeauthoryear {%
Bellman%
}{%
Bellman%
}{%
{\protect \APACyear {1957}}%
}]{%
bellman1957dynamic}
\APACinsertmetastar {%
bellman1957dynamic}%
\begin{APACrefauthors}%
Bellman, R.%
\end{APACrefauthors}%
\unskip\
\newblock
\APACrefYear{1957}.
\newblock
\APACrefbtitle {Dynamic programming} {Dynamic programming}.
\newblock
\APACaddressPublisher{}{Princeton University Press}.
\PrintBackRefs{\CurrentBib}

\bibitem [\protect \citeauthoryear {%
Botvinick%
}{%
Botvinick%
}{%
{\protect \APACyear {2012}}%
}]{%
botvinick2012hierarchical}
\APACinsertmetastar {%
botvinick2012hierarchical}%
\begin{APACrefauthors}%
Botvinick, M\BPBI M.%
\end{APACrefauthors}%
\unskip\
\newblock
\APACrefYearMonthDay{2012}{}{}.
\newblock
{\BBOQ}\APACrefatitle {Hierarchical reinforcement learning and decision making}
  {Hierarchical reinforcement learning and decision making}.{\BBCQ}
\newblock
\APACjournalVolNumPages{Current Opinion in Neurobiology}{22}{6}{956--962}.
\PrintBackRefs{\CurrentBib}

\bibitem [\protect \citeauthoryear {%
Botvinick%
, Niv%
\BCBL {}\ \BBA {} Barto%
}{%
Botvinick%
\ \protect \BOthers {.}}{%
{\protect \APACyear {2009}}%
}]{%
botvinick2009}
\APACinsertmetastar {%
botvinick2009}%
\begin{APACrefauthors}%
Botvinick, M\BPBI M.%
, Niv, Y.%
\BCBL {}\ \BBA {} Barto, A\BPBI C.%
\end{APACrefauthors}%
\unskip\
\newblock
\APACrefYearMonthDay{2009}{}{}.
\newblock
{\BBOQ}\APACrefatitle {Hierarchically organized behavior and its neural
  foundations: A reinforcement learning perspective} {Hierarchically organized
  behavior and its neural foundations: A reinforcement learning
  perspective}.{\BBCQ}
\newblock
\APACjournalVolNumPages{Cognition}{113}{3}{262--280}.
\PrintBackRefs{\CurrentBib}

\bibitem [\protect \citeauthoryear {%
Callaway%
\ \protect \BOthers {.}}{%
Callaway%
\ \protect \BOthers {.}}{%
{\protect \APACyear {2018}}%
}]{%
cogsci18-Callaway}
\APACinsertmetastar {%
cogsci18-Callaway}%
\begin{APACrefauthors}%
Callaway, F.%
, Lieder, F.%
, Das, P.%
, Gul, S.%
, Krueger, P.%
\BCBL {}\ \BBA {} Griffiths, T.%
\end{APACrefauthors}%
\unskip\
\newblock
\APACrefYearMonthDay{2018}{}{}.
\newblock
{\BBOQ}\APACrefatitle {A resource-rational analysis of human planning} {A
  resource-rational analysis of human planning}.{\BBCQ}
\newblock
\BIn{} \APACrefbtitle {{Proceedings of the Annual Conference of the Cognitive
  Science Society}.} {{Proceedings of the Annual Conference of the Cognitive
  Science Society}.}
\PrintBackRefs{\CurrentBib}

\bibitem [\protect \citeauthoryear {%
Collins%
\ \BBA {} Frank%
}{%
Collins%
\ \BBA {} Frank%
}{%
{\protect \APACyear {2013}}%
}]{%
Collins2013}
\APACinsertmetastar {%
Collins2013}%
\begin{APACrefauthors}%
Collins, A\BPBI G.%
\BCBT {}\ \BBA {} Frank, M\BPBI J.%
\end{APACrefauthors}%
\unskip\
\newblock
\APACrefYearMonthDay{2013}{}{}.
\newblock
{\BBOQ}\APACrefatitle {{Cognitive control over learning: Creating, clustering,
  and generalizing task-set structure}} {{Cognitive control over learning:
  Creating, clustering, and generalizing task-set structure}}.{\BBCQ}
\newblock
\APACjournalVolNumPages{Psychological Review}{120}{1}{190--229}.
\PrintBackRefs{\CurrentBib}

\bibitem [\protect \citeauthoryear {%
Cushman%
\ \BBA {} Morris%
}{%
Cushman%
\ \BBA {} Morris%
}{%
{\protect \APACyear {2015}}%
}]{%
Cushman13817}
\APACinsertmetastar {%
Cushman13817}%
\begin{APACrefauthors}%
Cushman, F.%
\BCBT {}\ \BBA {} Morris, A.%
\end{APACrefauthors}%
\unskip\
\newblock
\APACrefYearMonthDay{2015}{}{}.
\newblock
{\BBOQ}\APACrefatitle {Habitual control of goal selection in humans} {Habitual
  control of goal selection in humans}.{\BBCQ}
\newblock
\APACjournalVolNumPages{Proceedings of the National Academy of
  Sciences}{112}{45}{13817--13822}.
\PrintBackRefs{\CurrentBib}

\bibitem [\protect \citeauthoryear {%
Dietterich%
}{%
Dietterich%
}{%
{\protect \APACyear {2000}}%
}]{%
dietterich2000}
\APACinsertmetastar {%
dietterich2000}%
\begin{APACrefauthors}%
Dietterich, T\BPBI G.%
\end{APACrefauthors}%
\unskip\
\newblock
\APACrefYearMonthDay{2000}{}{}.
\newblock
{\BBOQ}\APACrefatitle {{Hierarchical Reinforcement Learning with the
  {M}{A}{X}{Q} Value Function Decomposition}} {{Hierarchical Reinforcement
  Learning with the {M}{A}{X}{Q} Value Function Decomposition}}.{\BBCQ}
\newblock
\APACjournalVolNumPages{Journal of artificial intelligence
  research}{13}{}{227--303}.
\PrintBackRefs{\CurrentBib}

\bibitem [\protect \citeauthoryear {%
Gershman%
, Horvitz%
\BCBL {}\ \BBA {} Tenenbaum%
}{%
Gershman%
\ \protect \BOthers {.}}{%
{\protect \APACyear {2015}}%
}]{%
Gershman2015computational}
\APACinsertmetastar {%
Gershman2015computational}%
\begin{APACrefauthors}%
Gershman, S\BPBI J.%
, Horvitz, E\BPBI J.%
\BCBL {}\ \BBA {} Tenenbaum, J\BPBI B.%
\end{APACrefauthors}%
\unskip\
\newblock
\APACrefYearMonthDay{2015}{}{}.
\newblock
{\BBOQ}\APACrefatitle {Computational rationality: A converging paradigm for
  intelligence in brains, minds, and machines} {Computational rationality: A
  converging paradigm for intelligence in brains, minds, and machines}.{\BBCQ}
\newblock
\APACjournalVolNumPages{Science}{349}{6245}{273--278}.
\PrintBackRefs{\CurrentBib}

\bibitem [\protect \citeauthoryear {%
Griffiths%
, Lieder%
\BCBL {}\ \BBA {} Goodman%
}{%
Griffiths%
\ \protect \BOthers {.}}{%
{\protect \APACyear {2015}}%
}]{%
Griffiths2015}
\APACinsertmetastar {%
Griffiths2015}%
\begin{APACrefauthors}%
Griffiths, T\BPBI L.%
, Lieder, F.%
\BCBL {}\ \BBA {} Goodman, N\BPBI D.%
\end{APACrefauthors}%
\unskip\
\newblock
\APACrefYearMonthDay{2015}{}{}.
\newblock
{\BBOQ}\APACrefatitle {{Rational Use of Cognitive Resources: Levels of Analysis
  Between the Computational and the Algorithmic}} {{Rational Use of Cognitive
  Resources: Levels of Analysis Between the Computational and the
  Algorithmic}}.{\BBCQ}
\newblock
\APACjournalVolNumPages{Topics in Cognitive Science}{7}{2}{217--229}.
\PrintBackRefs{\CurrentBib}

\bibitem [\protect \citeauthoryear {%
Haarnoja%
, Tang%
, Abbeel%
\BCBL {}\ \BBA {} Levine%
}{%
Haarnoja%
\ \protect \BOthers {.}}{%
{\protect \APACyear {2017}}%
}]{%
haarnoja2017}
\APACinsertmetastar {%
haarnoja2017}%
\begin{APACrefauthors}%
Haarnoja, T.%
, Tang, H.%
, Abbeel, P.%
\BCBL {}\ \BBA {} Levine, S.%
\end{APACrefauthors}%
\unskip\
\newblock
\APACrefYearMonthDay{2017}{}{}.
\newblock
{\BBOQ}\APACrefatitle {Reinforcement Learning with Deep Energy-Based Policies}
  {Reinforcement learning with deep energy-based policies}.{\BBCQ}
\newblock
\BIn{} \APACrefbtitle {{ICML}} {{ICML}}\ (\BPGS\ 1352--1361).
\PrintBackRefs{\CurrentBib}

\bibitem [\protect \citeauthoryear {%
Harb%
, Bacon%
, Klissarov%
\BCBL {}\ \BBA {} Precup%
}{%
Harb%
\ \protect \BOthers {.}}{%
{\protect \APACyear {2018}}%
}]{%
harb2018}
\APACinsertmetastar {%
harb2018}%
\begin{APACrefauthors}%
Harb, J.%
, Bacon, P\BHBI L.%
, Klissarov, M.%
\BCBL {}\ \BBA {} Precup, D.%
\end{APACrefauthors}%
\unskip\
\newblock
\APACrefYearMonthDay{2018}{}{}.
\newblock
{\BBOQ}\APACrefatitle {When waiting is not an option: Learning options with a
  deliberation cost} {When waiting is not an option: Learning options with a
  deliberation cost}.{\BBCQ}
\newblock
\BIn{} \APACrefbtitle {{Thirty-Second AAAI Conference on Artificial
  Intelligence}.} {{Thirty-Second AAAI Conference on Artificial Intelligence}.}
\PrintBackRefs{\CurrentBib}

\bibitem [\protect \citeauthoryear {%
{Hart}%
, {Nilsson}%
\BCBL {}\ \BBA {} {Raphael}%
}{%
{Hart}%
\ \protect \BOthers {.}}{%
{\protect \APACyear {1968}}%
}]{%
hart1968formal}
\APACinsertmetastar {%
hart1968formal}%
\begin{APACrefauthors}%
{Hart}, P\BPBI E.%
, {Nilsson}, N\BPBI J.%
\BCBL {}\ \BBA {} {Raphael}, B.%
\end{APACrefauthors}%
\unskip\
\newblock
\APACrefYearMonthDay{1968}{}{}.
\newblock
{\BBOQ}\APACrefatitle {A Formal Basis for the Heuristic Determination of
  Minimum Cost Paths} {A formal basis for the heuristic determination of
  minimum cost paths}.{\BBCQ}
\newblock
\APACjournalVolNumPages{IEEE Transactions on Systems Science and
  Cybernetics}{4}{2}{100-107}.
\PrintBackRefs{\CurrentBib}

\bibitem [\protect \citeauthoryear {%
Ho%
, Abel%
, Cohen%
, Littman%
\BCBL {}\ \BBA {} Griffiths%
}{%
Ho%
\ \protect \BOthers {.}}{%
{\protect \APACyear {2020}}%
}]{%
ho2020}
\APACinsertmetastar {%
ho2020}%
\begin{APACrefauthors}%
Ho, M\BPBI K.%
, Abel, D.%
, Cohen, J.%
, Littman, M\BPBI L.%
\BCBL {}\ \BBA {} Griffiths, T\BPBI L.%
\end{APACrefauthors}%
\unskip\
\newblock
\APACrefYearMonthDay{2020}{}{}.
\newblock
{\BBOQ}\APACrefatitle {{The Efficiency of Human Cognition Reflects Planned
  Information Processing}} {{The Efficiency of Human Cognition Reflects Planned
  Information Processing}}.{\BBCQ}
\newblock
\BIn{} \APACrefbtitle {{Thirty-Fourth AAAI Conference on Artificial
  Intelligence}.} {{Thirty-Fourth AAAI Conference on Artificial Intelligence}.}
\PrintBackRefs{\CurrentBib}

\bibitem [\protect \citeauthoryear {%
Ho%
, Abel%
, Griffiths%
\BCBL {}\ \BBA {} Littman%
}{%
Ho%
\ \protect \BOthers {.}}{%
{\protect \APACyear {2019}}%
}]{%
ho2019}
\APACinsertmetastar {%
ho2019}%
\begin{APACrefauthors}%
Ho, M\BPBI K.%
, Abel, D.%
, Griffiths, T\BPBI L.%
\BCBL {}\ \BBA {} Littman, M\BPBI L.%
\end{APACrefauthors}%
\unskip\
\newblock
\APACrefYearMonthDay{2019}{}{}.
\newblock
{\BBOQ}\APACrefatitle {The value of abstraction} {The value of
  abstraction}.{\BBCQ}
\newblock
\APACjournalVolNumPages{Current Opinion in Behavioral
  Sciences}{29}{}{111--116}.
\PrintBackRefs{\CurrentBib}

\bibitem [\protect \citeauthoryear {%
Huys%
\ \protect \BOthers {.}}{%
Huys%
\ \protect \BOthers {.}}{%
{\protect \APACyear {2012}}%
}]{%
Huys2012}
\APACinsertmetastar {%
Huys2012}%
\begin{APACrefauthors}%
Huys, Q\BPBI J\BPBI M.%
, Eshel, N.%
, O'Nions, E.%
, Sheridan, L.%
, Dayan, P.%
\BCBL {}\ \BBA {} Roiser, J\BPBI P.%
\end{APACrefauthors}%
\unskip\
\newblock
\APACrefYearMonthDay{2012}{}{}.
\newblock
{\BBOQ}\APACrefatitle {{Bonsai trees in your head: how the Pavlovian system
  sculpts goal-directed choices by pruning decision trees}} {{Bonsai trees in
  your head: how the Pavlovian system sculpts goal-directed choices by pruning
  decision trees}}.{\BBCQ}
\newblock
\APACjournalVolNumPages{PLoS Computational Biology}{8}{3}{e1002410}.
\PrintBackRefs{\CurrentBib}

\bibitem [\protect \citeauthoryear {%
Huys%
\ \protect \BOthers {.}}{%
Huys%
\ \protect \BOthers {.}}{%
{\protect \APACyear {2015}}%
}]{%
Huys2015}
\APACinsertmetastar {%
Huys2015}%
\begin{APACrefauthors}%
Huys, Q\BPBI J\BPBI M.%
, Lally, N.%
, Faulkner, P.%
, Eshel, N.%
, Seifritz, E.%
, Gershman, S\BPBI J.%
\BDBL {}Roiser, J\BPBI P.%
\end{APACrefauthors}%
\unskip\
\newblock
\APACrefYearMonthDay{2015}{}{}.
\newblock
{\BBOQ}\APACrefatitle {{Interplay of approximate planning strategies.}}
  {{Interplay of approximate planning strategies.}}{\BBCQ}
\newblock
\APACjournalVolNumPages{Proceedings of the National Academy of
  Sciences}{112}{10}{3098--103}.
\PrintBackRefs{\CurrentBib}

\bibitem [\protect \citeauthoryear {%
Jinnai%
, Abel%
, Hershkowitz%
, Littman%
\BCBL {}\ \BBA {} Konidaris%
}{%
Jinnai%
\ \protect \BOthers {.}}{%
{\protect \APACyear {2019}}%
}]{%
jinnai2018}
\APACinsertmetastar {%
jinnai2018}%
\begin{APACrefauthors}%
Jinnai, Y.%
, Abel, D.%
, Hershkowitz, D.%
, Littman, M.%
\BCBL {}\ \BBA {} Konidaris, G.%
\end{APACrefauthors}%
\unskip\
\newblock
\APACrefYearMonthDay{2019}{}{}.
\newblock
{\BBOQ}\APACrefatitle {Finding Options that Minimize Planning Time} {Finding
  options that minimize planning time}.{\BBCQ}
\newblock
\BIn{} \APACrefbtitle {{ICML}} {{ICML}}\ (\BVOL~97, \BPGS\ 3120--3129).
\PrintBackRefs{\CurrentBib}

\bibitem [\protect \citeauthoryear {%
Kaplan%
\ \BBA {} Simon%
}{%
Kaplan%
\ \BBA {} Simon%
}{%
{\protect \APACyear {1990}}%
}]{%
kaplan1990search}
\APACinsertmetastar {%
kaplan1990search}%
\begin{APACrefauthors}%
Kaplan, C\BPBI A.%
\BCBT {}\ \BBA {} Simon, H\BPBI A.%
\end{APACrefauthors}%
\unskip\
\newblock
\APACrefYearMonthDay{1990}{}{}.
\newblock
{\BBOQ}\APACrefatitle {In search of insight} {In search of insight}.{\BBCQ}
\newblock
\APACjournalVolNumPages{Cognitive Psychology}{22}{3}{374 - 419}.
\PrintBackRefs{\CurrentBib}

\bibitem [\protect \citeauthoryear {%
Keramati%
, Smittenaar%
, Dolan%
\BCBL {}\ \BBA {} Dayan%
}{%
Keramati%
\ \protect \BOthers {.}}{%
{\protect \APACyear {2016}}%
}]{%
keramati2016}
\APACinsertmetastar {%
keramati2016}%
\begin{APACrefauthors}%
Keramati, M.%
, Smittenaar, P.%
, Dolan, R\BPBI J.%
\BCBL {}\ \BBA {} Dayan, P.%
\end{APACrefauthors}%
\unskip\
\newblock
\APACrefYearMonthDay{2016}{}{}.
\newblock
{\BBOQ}\APACrefatitle {Adaptive integration of habits into depth-limited
  planning defines a habitual-goal--directed spectrum} {Adaptive integration of
  habits into depth-limited planning defines a habitual-goal--directed
  spectrum}.{\BBCQ}
\newblock
\APACjournalVolNumPages{Proceedings of the National Academy of
  Sciences}{113}{45}{12868--12873}.
\PrintBackRefs{\CurrentBib}

\bibitem [\protect \citeauthoryear {%
Lewis%
, Howes%
\BCBL {}\ \BBA {} Singh%
}{%
Lewis%
\ \protect \BOthers {.}}{%
{\protect \APACyear {2014}}%
}]{%
lewis2014computational}
\APACinsertmetastar {%
lewis2014computational}%
\begin{APACrefauthors}%
Lewis, R\BPBI L.%
, Howes, A.%
\BCBL {}\ \BBA {} Singh, S.%
\end{APACrefauthors}%
\unskip\
\newblock
\APACrefYearMonthDay{2014}{}{}.
\newblock
{\BBOQ}\APACrefatitle {Computational Rationality: Linking Mechanism and
  Behavior Through Bounded Utility Maximization} {Computational rationality:
  Linking mechanism and behavior through bounded utility maximization}.{\BBCQ}
\newblock
\APACjournalVolNumPages{Topics in Cognitive Science}{6}{2}{279-311}.
\PrintBackRefs{\CurrentBib}

\bibitem [\protect \citeauthoryear {%
Lieder%
\ \BBA {} Griffiths%
}{%
Lieder%
\ \BBA {} Griffiths%
}{%
{\protect \APACyear {2020}}%
}]{%
Lieder2020}
\APACinsertmetastar {%
Lieder2020}%
\begin{APACrefauthors}%
Lieder, F.%
\BCBT {}\ \BBA {} Griffiths, T\BPBI L.%
\end{APACrefauthors}%
\unskip\
\newblock
\APACrefYearMonthDay{2020}{}{}.
\newblock
{\BBOQ}\APACrefatitle {Resource-rational analysis: understanding human
  cognition as the optimal use of limited computational resources}
  {Resource-rational analysis: understanding human cognition as the optimal use
  of limited computational resources}.{\BBCQ}
\newblock
\APACjournalVolNumPages{Behavioral and Brain Sciences}{}{}{1--60}.
\PrintBackRefs{\CurrentBib}

\bibitem [\protect \citeauthoryear {%
MacGregor%
, Ormerod%
\BCBL {}\ \BBA {} Chronicle%
}{%
MacGregor%
\ \protect \BOthers {.}}{%
{\protect \APACyear {2001}}%
}]{%
MacGregor2001}
\APACinsertmetastar {%
MacGregor2001}%
\begin{APACrefauthors}%
MacGregor, J\BPBI N.%
, Ormerod, T\BPBI C.%
\BCBL {}\ \BBA {} Chronicle, E\BPBI P.%
\end{APACrefauthors}%
\unskip\
\newblock
\APACrefYearMonthDay{2001}{}{}.
\newblock
{\BBOQ}\APACrefatitle {Information processing and insight: a process model of
  performance on the nine-dot and related problems.} {Information processing
  and insight: a process model of performance on the nine-dot and related
  problems.}{\BBCQ}
\newblock
\APACjournalVolNumPages{Journal of Experimental Psychology: Learning, Memory,
  and Cognition}{27}{1}{176}.
\PrintBackRefs{\CurrentBib}

\bibitem [\protect \citeauthoryear {%
Maisto%
, Donnarumma%
\BCBL {}\ \BBA {} Pezzulo%
}{%
Maisto%
\ \protect \BOthers {.}}{%
{\protect \APACyear {2015}}%
}]{%
Maisto2015}
\APACinsertmetastar {%
Maisto2015}%
\begin{APACrefauthors}%
Maisto, D.%
, Donnarumma, F.%
\BCBL {}\ \BBA {} Pezzulo, G.%
\end{APACrefauthors}%
\unskip\
\newblock
\APACrefYearMonthDay{2015}{}{}.
\newblock
{\BBOQ}\APACrefatitle {{Divide et impera: subgoaling reduces the complexity of
  probabilistic inference and problem solving}} {{Divide et impera: subgoaling
  reduces the complexity of probabilistic inference and problem
  solving}}.{\BBCQ}
\newblock
\APACjournalVolNumPages{Journal of The Royal Society
  Interface}{12}{104}{20141335--20141335}.
\PrintBackRefs{\CurrentBib}

\bibitem [\protect \citeauthoryear {%
Newell%
\ \BBA {} Simon%
}{%
Newell%
\ \BBA {} Simon%
}{%
{\protect \APACyear {1972}}%
}]{%
NewellSimon1972}
\APACinsertmetastar {%
NewellSimon1972}%
\begin{APACrefauthors}%
Newell, A.%
\BCBT {}\ \BBA {} Simon, H\BPBI A.%
\end{APACrefauthors}%
\unskip\
\newblock
\APACrefYear{1972}.
\newblock
\APACrefbtitle {Human problem solving} {Human problem solving}.
\newblock
\APACaddressPublisher{Englewood Cliffs, NJ}{Prentice-Hall}.
\PrintBackRefs{\CurrentBib}

\bibitem [\protect \citeauthoryear {%
Nilsson%
}{%
Nilsson%
}{%
{\protect \APACyear {1971}}%
}]{%
nilsson1971}
\APACinsertmetastar {%
nilsson1971}%
\begin{APACrefauthors}%
Nilsson, N\BPBI J.%
\end{APACrefauthors}%
\unskip\
\newblock
\APACrefYear{1971}.
\newblock
\APACrefbtitle {Problem-solving methods in artificial intelligence}
  {Problem-solving methods in artificial intelligence}.
\newblock
\APACaddressPublisher{}{McGraw-Hill}.
\PrintBackRefs{\CurrentBib}

\bibitem [\protect \citeauthoryear {%
Parr%
\ \BBA {} Russell%
}{%
Parr%
\ \BBA {} Russell%
}{%
{\protect \APACyear {1998}}%
}]{%
parr1998}
\APACinsertmetastar {%
parr1998}%
\begin{APACrefauthors}%
Parr, R.%
\BCBT {}\ \BBA {} Russell, S\BPBI J.%
\end{APACrefauthors}%
\unskip\
\newblock
\APACrefYearMonthDay{1998}{}{}.
\newblock
{\BBOQ}\APACrefatitle {Reinforcement learning with hierarchies of machines}
  {Reinforcement learning with hierarchies of machines}.{\BBCQ}
\newblock
\BIn{} \APACrefbtitle {{Advances in Neural Information Processing Systems}}
  {{Advances in Neural Information Processing Systems}}\ (\BPGS\ 1043--1049).
\PrintBackRefs{\CurrentBib}

\bibitem [\protect \citeauthoryear {%
Puterman%
}{%
Puterman%
}{%
{\protect \APACyear {1994}}%
}]{%
puterman1994markov}
\APACinsertmetastar {%
puterman1994markov}%
\begin{APACrefauthors}%
Puterman, M\BPBI L.%
\end{APACrefauthors}%
\unskip\
\newblock
\APACrefYear{1994}.
\newblock
\APACrefbtitle {Markov Decision Processes: Discrete Stochastic Dynamic
  Programming} {Markov decision processes: Discrete stochastic dynamic
  programming}.
\newblock
\APACaddressPublisher{}{John Wiley \& Sons, Inc.}
\PrintBackRefs{\CurrentBib}

\bibitem [\protect \citeauthoryear {%
Ribas-Fernandes%
\ \protect \BOthers {.}}{%
Ribas-Fernandes%
\ \protect \BOthers {.}}{%
{\protect \APACyear {2011}}%
}]{%
ribasfernandes2011neural}
\APACinsertmetastar {%
ribasfernandes2011neural}%
\begin{APACrefauthors}%
Ribas-Fernandes, J.%
, Solway, A.%
, Diuk, C.%
, McGuire, J.%
, Barto, A.%
, Niv, Y.%
\BCBL {}\ \BBA {} Botvinick, M.%
\end{APACrefauthors}%
\unskip\
\newblock
\APACrefYearMonthDay{2011}{}{}.
\newblock
{\BBOQ}\APACrefatitle {A Neural Signature of Hierarchical Reinforcement
  Learning} {A neural signature of hierarchical reinforcement learning}.{\BBCQ}
\newblock
\APACjournalVolNumPages{Neuron}{71}{2}{370 - 379}.
\PrintBackRefs{\CurrentBib}

\bibitem [\protect \citeauthoryear {%
Russell%
\ \BBA {} Norvig%
}{%
Russell%
\ \BBA {} Norvig%
}{%
{\protect \APACyear {2009}}%
}]{%
russell2009artificial}
\APACinsertmetastar {%
russell2009artificial}%
\begin{APACrefauthors}%
Russell, S.%
\BCBT {}\ \BBA {} Norvig, P.%
\end{APACrefauthors}%
\unskip\
\newblock
\APACrefYear{2009}.
\newblock
\APACrefbtitle {Artificial Intelligence: A Modern Approach} {Artificial
  intelligence: A modern approach}\ (\PrintOrdinal{3rd}\ \BEd).
\newblock
\APACaddressPublisher{USA}{Prentice Hall Press}.
\PrintBackRefs{\CurrentBib}

\bibitem [\protect \citeauthoryear {%
Sacerdoti%
}{%
Sacerdoti%
}{%
{\protect \APACyear {1974}}%
}]{%
sacerdoti1974planning}
\APACinsertmetastar {%
sacerdoti1974planning}%
\begin{APACrefauthors}%
Sacerdoti, E\BPBI D.%
\end{APACrefauthors}%
\unskip\
\newblock
\APACrefYearMonthDay{1974}{}{}.
\newblock
{\BBOQ}\APACrefatitle {Planning in a hierarchy of abstraction spaces} {Planning
  in a hierarchy of abstraction spaces}.{\BBCQ}
\newblock
\APACjournalVolNumPages{{Artificial Intelligence}}{5}{2}{115--135}.
\PrintBackRefs{\CurrentBib}

\bibitem [\protect \citeauthoryear {%
Schapiro%
, Rogers%
, Cordova%
, Turk-Browne%
\BCBL {}\ \BBA {} Botvinick%
}{%
Schapiro%
\ \protect \BOthers {.}}{%
{\protect \APACyear {2013}}%
}]{%
Schapiro2013}
\APACinsertmetastar {%
Schapiro2013}%
\begin{APACrefauthors}%
Schapiro, A\BPBI C.%
, Rogers, T\BPBI T.%
, Cordova, N\BPBI I.%
, Turk-Browne, N\BPBI B.%
\BCBL {}\ \BBA {} Botvinick, M\BPBI M.%
\end{APACrefauthors}%
\unskip\
\newblock
\APACrefYearMonthDay{2013}{}{}.
\newblock
{\BBOQ}\APACrefatitle {{Neural representations of events arise from temporal
  community structure}} {{Neural representations of events arise from temporal
  community structure}}.{\BBCQ}
\newblock
\APACjournalVolNumPages{Nature Neuroscience}{16}{4}{486--492}.
\PrintBackRefs{\CurrentBib}

\bibitem [\protect \citeauthoryear {%
{\c{S}}im{\c{s}}ek%
\ \BBA {} Barto%
}{%
{\c{S}}im{\c{s}}ek%
\ \BBA {} Barto%
}{%
{\protect \APACyear {2009}}%
}]{%
csimcsek2009}
\APACinsertmetastar {%
csimcsek2009}%
\begin{APACrefauthors}%
{\c{S}}im{\c{s}}ek, {\"O}.%
\BCBT {}\ \BBA {} Barto, A\BPBI G.%
\end{APACrefauthors}%
\unskip\
\newblock
\APACrefYearMonthDay{2009}{}{}.
\newblock
{\BBOQ}\APACrefatitle {Skill characterization based on betweenness} {Skill
  characterization based on betweenness}.{\BBCQ}
\newblock
\BIn{} \APACrefbtitle {{Advances in Neural Information Processing Systems}}
  {{Advances in Neural Information Processing Systems}}\ (\BPGS\ 1497--1504).
\PrintBackRefs{\CurrentBib}

\bibitem [\protect \citeauthoryear {%
Solway%
\ \protect \BOthers {.}}{%
Solway%
\ \protect \BOthers {.}}{%
{\protect \APACyear {2014}}%
}]{%
solway2014}
\APACinsertmetastar {%
solway2014}%
\begin{APACrefauthors}%
Solway, A.%
, Diuk, C.%
, C{\'o}rdova, N.%
, Yee, D.%
, Barto, A\BPBI G.%
, Niv, Y.%
\BCBL {}\ \BBA {} Botvinick, M\BPBI M.%
\end{APACrefauthors}%
\unskip\
\newblock
\APACrefYearMonthDay{2014}{}{}.
\newblock
{\BBOQ}\APACrefatitle {Optimal behavioral hierarchy} {Optimal behavioral
  hierarchy}.{\BBCQ}
\newblock
\APACjournalVolNumPages{PLoS Computational Biology}{10}{8}{e1003779}.
\PrintBackRefs{\CurrentBib}

\bibitem [\protect \citeauthoryear {%
Sutton%
, Precup%
\BCBL {}\ \BBA {} Singh%
}{%
Sutton%
\ \protect \BOthers {.}}{%
{\protect \APACyear {1999}}%
}]{%
sutton1999}
\APACinsertmetastar {%
sutton1999}%
\begin{APACrefauthors}%
Sutton, R\BPBI S.%
, Precup, D.%
\BCBL {}\ \BBA {} Singh, S.%
\end{APACrefauthors}%
\unskip\
\newblock
\APACrefYearMonthDay{1999}{}{}.
\newblock
{\BBOQ}\APACrefatitle {Between MDPs and semi-MDPs: A framework for temporal
  abstraction in reinforcement learning} {Between mdps and semi-mdps: A
  framework for temporal abstraction in reinforcement learning}.{\BBCQ}
\newblock
\APACjournalVolNumPages{{Artificial Intelligence}}{112}{1-2}{181--211}.
\PrintBackRefs{\CurrentBib}

\bibitem [\protect \citeauthoryear {%
Tomov%
, Yagati%
, Kumar%
, Yang%
\BCBL {}\ \BBA {} Gershman%
}{%
Tomov%
\ \protect \BOthers {.}}{%
{\protect \APACyear {2020}}%
}]{%
Tomov2020}
\APACinsertmetastar {%
Tomov2020}%
\begin{APACrefauthors}%
Tomov, M\BPBI S.%
, Yagati, S.%
, Kumar, A.%
, Yang, W.%
\BCBL {}\ \BBA {} Gershman, S\BPBI J.%
\end{APACrefauthors}%
\unskip\
\newblock
\APACrefYearMonthDay{2020}{}{}.
\newblock
{\BBOQ}\APACrefatitle {Discovery of hierarchical representations for efficient
  planning} {Discovery of hierarchical representations for efficient
  planning}.{\BBCQ}
\newblock
\APACjournalVolNumPages{PLOS Computational Biology}{16}{4}{}.
\PrintBackRefs{\CurrentBib}

\bibitem [\protect \citeauthoryear {%
van Opheusden%
, Galbiati%
, Bnaya%
, Li%
\BCBL {}\ \BBA {} Ma%
}{%
van Opheusden%
\ \protect \BOthers {.}}{%
{\protect \APACyear {2017}}%
}]{%
vanopheusden2017}
\APACinsertmetastar {%
vanopheusden2017}%
\begin{APACrefauthors}%
van Opheusden, B.%
, Galbiati, G.%
, Bnaya, Z.%
, Li, Y.%
\BCBL {}\ \BBA {} Ma, W\BPBI J.%
\end{APACrefauthors}%
\unskip\
\newblock
\APACrefYearMonthDay{2017}{}{}.
\newblock
{\BBOQ}\APACrefatitle {A computational model for decision tree search.} {A
  computational model for decision tree search.}{\BBCQ}
\newblock
\BIn{} \APACrefbtitle {{Proceedings of the Annual Conference of the Cognitive
  Science Society}.} {{Proceedings of the Annual Conference of the Cognitive
  Science Society}.}
\PrintBackRefs{\CurrentBib}

\end{thebibliography}

\end{document}